\documentclass[10pt,conference]{IEEEtran}
\IEEEoverridecommandlockouts

\usepackage{cite}
\usepackage{amsmath,amssymb,amsfonts}
\usepackage{algorithmic}
\usepackage{graphicx}
\usepackage{textcomp}
\usepackage{xcolor}
\usepackage{multirow}
\usepackage{booktabs}
\usepackage{soul}
\def\BibTeX{{\rm B\kern-.05em{\sc i\kern-.025em b}\kern-.08em
    T\kern-.1667em\lower.7ex\hbox{E}\kern-.125emX}}
\begin{document}

\title{Multimodal Machine Learning Can Predict\\
Videoconference Fluidity and Enjoyment\\ 

\thanks{A.C., D.P., and D.F. are supported by NYU Discovery Research Fund for Human Health. A.C. is supported by National Research Service Award, NIDCD/NIH (F32DC018205) and Leon Levy Scholarships in Neuroscience, Leon Levy Foundation and New York Academy of Sciences. The funders have no role in study design, data collection and analysis, decision to publish, or preparation of the manuscript. This work was supported in part through the NYU IT High Performance Computing resources, services, and staff expertise. 

© 2025 IEEE.  Personal use of this material is permitted. Permission from IEEE must be obtained for all other uses, in any current or future media, including reprinting/republishing this material for advertising or promotional purposes, creating new collective works, for resale or redistribution to servers or lists, or reuse of any copyrighted component of this work in other works.}
}

\author{\IEEEauthorblockN{Andrew Chang\IEEEauthorrefmark{1}, Viswadruth Akkaraju\IEEEauthorrefmark{1}, Ray McFadden Cogliano\IEEEauthorrefmark{1}, David Poeppel\IEEEauthorrefmark{1}\IEEEauthorrefmark{2}, Dustin Freeman\IEEEauthorrefmark{1}}
\IEEEauthorblockA{\IEEEauthorrefmark{1}New York University, New York, USA}
\IEEEauthorblockA{\IEEEauthorrefmark{2}Max Planck Society, Munich, Germany}
}

\maketitle

\begin{abstract}

Videoconferencing is now a frequent mode of communication in both professional and informal settings, yet it often lacks the fluidity and enjoyment of in-person conversation. This study leverages multimodal machine learning to predict moments of negative experience in videoconferencing. We sampled thousands of short clips from the RoomReader corpus \cite{reverdy-etal-2022-roomreader}, extracting audio embeddings, facial actions, and body motion features to train models for identifying low conversational fluidity, low enjoyment, and classifying conversational events (backchanneling, interruption, or gap). Our best models achieved an ROC-AUC of up to 0.87 on hold-out videoconference sessions, with domain-general audio features proving most critical. This work demonstrates that multimodal audio-video signals can effectively predict high-level subjective conversational outcomes. In addition, this is a contribution to research on videoconferencing user experience by showing that multimodal machine learning can be used to identify rare moments of negative user experience for further study or mitigation.

\end{abstract}

\begin{IEEEkeywords}
multimodal features, videoconferencing, subjective rating, conversational speech, facial action
\end{IEEEkeywords}

\section{Introduction and Goals}

Videoconferencing, as a medium, attempts to simulate in-person conversation. However, it often results in conversations that are less effective for fostering genuine interaction and intimacy, and ``Zoom fatigue'' is one of the well-known negative experience outcomes \cite{bennett2021videoconference, NADLER2020102613 }. The causes of negative experience remain unclear and may involve factors such as signal delay/latency, reduction in signal quality due to compression, packet loss, and mismatch of perceptual cues.

The bulk of time and data in any videoconferencing session is unproblematic and thus uninteresting for failure analysis. It is more useful to find specific moments of negative experience. Even if these events are infrequent, their negative impact on the conversation's participants is disproportionate \cite{powers2011effect}. However, the definition of negative psychological experience is a high-level subjective outcome that usually requires manual, time-consuming annotation. %

\begin{figure}[htbp]
\vspace{-5pt}
\centerline{\includegraphics[width=0.5\textwidth]{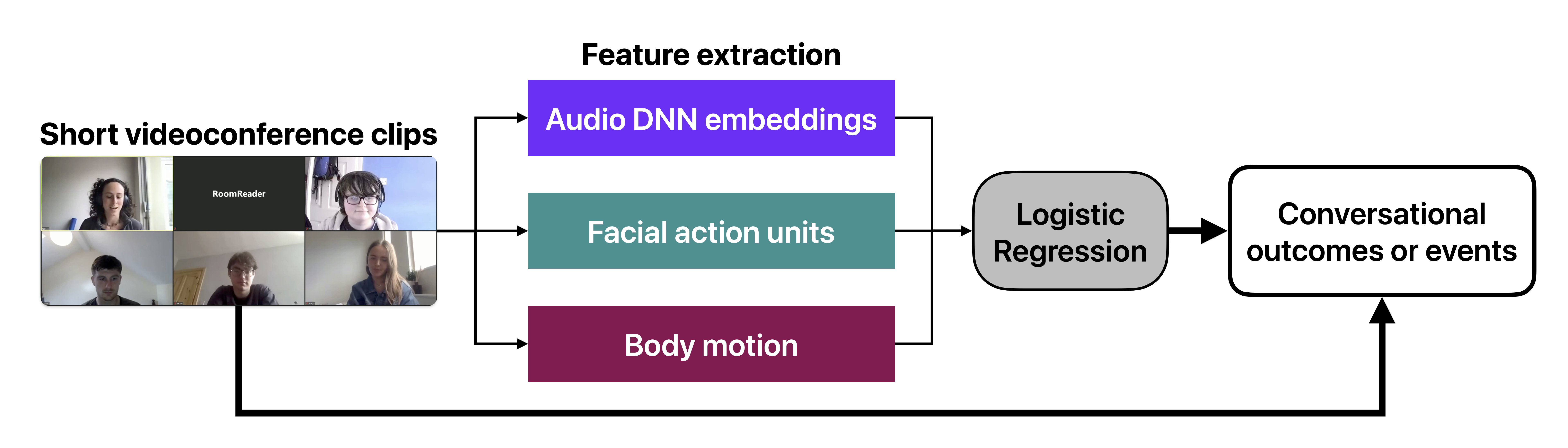}}
\vspace{-5pt} 
\caption{The flow diagram of the multimodal machine learning}
\label{model_dataflow}
\vspace{-10pt} 
\end{figure}

We aimed to determine how well a machine learning model trained on multimodal signals (audio, facial actions, body motion) can predict subjective conversational outcomes (i.e., fluidity and enjoyment) or classify conversational events (backchanneling, interruption, gap) in a short videoconference period (Fig. \ref{model_dataflow}). Specifically, we examined: (1) the contribution of each feature domain to model performance, (2) whether the pre-event features can also predict outcomes, (3) the relationship between low-level fluidity and high-level enjoyment, and (4) the model's ability to differentiate conversational events. To ensure generalizability, the trained model was tested on new videoconference sessions.

\section{Prior Work}
Most prior research on applying machine learning to videoconferencing has pursued goals different from ours, focusing primarily on extending video analytics techniques to dynamic, real-world multiparty videoconferencing scenarios \cite{vishruth2024videoconference}, or facilitating signal quality \cite{rao2021conferencingspeech}. Some studies were similar to those of the current work: one study used multimodal machine learning, trained on facial expressions, audio, eye gaze and more, predicted collaborative problem solving performance via videoconferencing \cite{subburaj2020multimodal, stewart2021multimodal}. Another study with a similar design showed that the combined video, speech and body motion features are statistically associated with subjective perception of the collaboration \cite{vrzakova2020focused}. A recent study showed that nonverbal features can predict the next speaker in multiparty videoconference \cite{mizuno2023next}. Another recent study experimentally manipulated different videoconference system parameters (signal delay, jitter, and packet loss rate), and demonstrated that a machine model trained on facial and speech features can predict participants' quality of experience \cite{Bingol2024}.

Our study distinguishes itself from previous research in several key ways: (1) it leverages a significantly larger dataset ($\sim$3000 vs. dozens to hundreds of sessions), (2) it models videoconferencing experiences within a much shorter timescale (7 s vs. $\sim$15--60 mins), and (3) it focuses on high-level psychological and emotional dimensions of social interaction.

\section{Dataset}
\subsection{Video Clips}
\subsubsection{Videoconference Corpus: RoomReader}
We extracted the videoconference clips from the RoomReader corpus \cite{reverdy-etal-2022-roomreader}. This corpus contains 30 Zoom videoconference sessions, each has 4 or 5 participants and last 8-30 mins, totalling near 9 hours and 118 participants (91 English native speakers). The conversations were mostly on collaborative quiz games and icebreakers. The procedure of the study was carefully designed to elicit naturalistic, spontaneous speech, i.e. turn-taking among participants based on their own initiative, without an explicit turn order. 

\subsubsection{Clip Selection}
We aimed to reduce the clip count to a manageable size for rating and increase the representation of negative moments in model training. For this work, we chose to focus on changes in turn-taking, where either nobody was speaking or two or more people were speaking simultaneously.

We implemented this selection according to the following rules: We set root-mean-square 0.05 as a binary threshold for detecting each speaker's audio. When all speaker’s audio was below threshold for 0.75 s (much longer than the typical turn-taking transitional gap of 0.2 s in-person or 0.5 s on videoconference \cite{boland2022zoom}), we marked that as the beginning of a silence period. When the number of people speaking simultaneously exceeded one, we marked that as the beginning of an overlap period. A total of 1,508 marked time point in each category (3,016 clips in total) were sampled in the current study. For each marked time point, we extracted a group Zoom video clip from 3 s before to 4 s after. We did not highlight any information about how it was marked, or which speakers were involved to the raters. To help raters identify the moment of interest, we added a red border around each video clip from the marked time until 0.5 s after.

\subsection{Clip Rating Survey}

\subsubsection{Raters}
The raters were students at New York University who signed up for the studies via an online platform and received course credit for completing the experiments. In total, 528 raters (274 women, 167 men, 87 non-reporting or other) completed the survey, with ages ranging from 18 to 36 (median 20), and 242 (46\%) identified as native English speakers. They reported a median of 4 hours of weekly video chat use, with time divided between friends and family (60\%), study or work (12\%), and classes or conferences (18\%), showing a balance between informal and formal settings.

\subsubsection{Survey Procedure}
Participants completed the survey on the Qualtrics platform using their own computers, wearing headphones in a quiet environment. Each was assigned 120 randomly sampled clips to rate, taking under an hour in total, with the option to pause and resume within a week. A reliability block of 8 identical clips for all participants was presented at the beginning and end of the survey, while the remaining 104 clips were unique to each participant. For each clip, participants rated it on a 5-point Likert scale (1: low, 5: high) based on the following questions: (1) "How \textbf{fluid} is this conversation after the red frame appears?" (2) "How much did the group \textbf{enjoy} the conversation?". Also, a multiple choice question: (3) "What happens in the video when the red frame appears?" (\textbf{event}) (a) someone attempts to \textbf{interrupt} the speaker (which succeeds or fails); (b) \textbf{backchanneling} (laughing, agreeing, “uh-huh”); (c) a \textbf{gap} in conversation, where no one makes noise; (d) a sound unrelated to the conversation: non-speech voice (e.g. coughing, laughing), noises in environment (door closing, typing); (e) none of the above. Note we did not give participants a specific definition of 'fluid' intentionally, as our pilot focus group did not report any confusion. After all, it is possible that an overly-complex definition would not even be read by survey participants. Instead we said: \textit{"We want your help to judge whether, at the time point of the red frame, the conversation feels 'wrong,' 'non-fluid,' unnatural, or uncomfortable, based on your own instincts…just track your feelings after watching the section of conversation."}

\subsubsection{Data Preprocessing, Reliability and Inclusion}

\begin{figure}
\centerline{\includegraphics[width=0.45\textwidth]{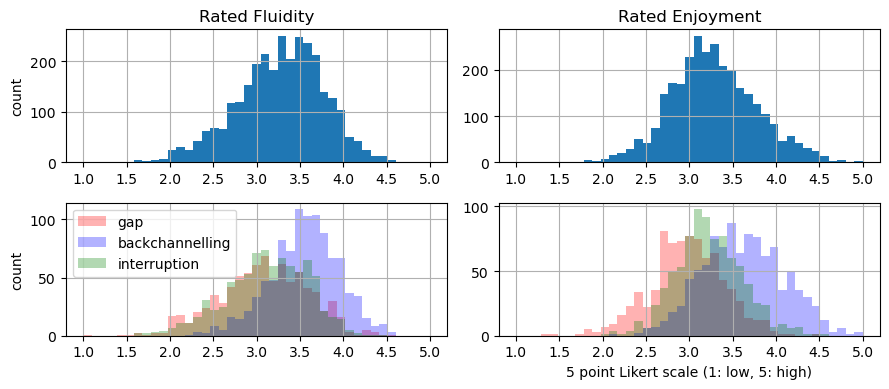}}
\vspace{-10pt} 
\caption{Mean Ratings of Fluidity and Enjoyment}
\label{response_hist_compare}
\vspace{-10pt}
\end{figure}

In any large-scale online survey, some participants will rush through without answering thoughtfully. We used reliability blocks to identify and exclude unreliable raters. For each rater, we calculated Pearson's \textit{r} between their Fluidity rating of 16 reliability clips and the average rating from other participants. Data from 350 raters with \textit{r} above 0.2 were included in further analysis. For each clip, we averaged ratings across reliable raters. Clips rated by fewer than 4 raters were excluded, leaving 2,992 clips for analyses. For event classification, each clip with the most frequently (also exceeding 40\%) annotated event  was included in the analyses, resulting in 810 to 996 clips per event. See Fig. \ref{response_hist_compare} for distributions.

\subsubsection{Exploratory Analyses of Conversational Events}

We explored whether the differently annotated conversational events are associated with rated Fluidity and Enjoyment levels (Fig. \ref{response_hist_compare}). Regarding Fluidity, the backchanneling clips were rated higher than interruption (\textit{t}(1841) = 21.5, \textit{p} $<$ .001) or gap (\textit{t}(1806) = 21.8, \textit{p} $<$ .001), but no difference between interruption and gap (\textit{t}(1653) = 1.5, \textit{p} = .129). Regarding Enjoyment, the backchanneling clips were rated higher than interruption (\textit{t}(1841) = 19.5, \textit{p} $<$ .001) or gap (\textit{t}(1806) = 30.3, \textit{p} $<$ .001), and interruption clips were higher than gap (\textit{t}(1653) = 12.5, \textit{p} $<$ .001). These interesting findings suggest that: (1) backchanneling is considered positive from both perspectives; (2) although acoustically distinct, both gaps and interruptions are equally detrimental to conversational fluidity; (3) despite both involving overlapping speakers' voices, backchanneling and interruptions are functionally very different; (4) gaps are worse than interruptions for overall enjoyment, implying that awkwardness is more damaging than rudeness.

\subsubsection{Binary Data Transformation}
We converted the Likert scale into binary for use in machine learning modeling, since we are primarily interested in the most severe types of conversational failure. The ultimate goal of this study is to identify events with negative user experiences, so modeling differences between events rated 3 (mediocre) or above was not a priority. Instead of asking participants to make binary judgments, we used a Likert scale to obtain an interrater reliability measure and to understand the nature of scale use. We set a threshold of 2.5 for both scales, as visual inspection revealed a noticeable step in the distributions (Fig. \ref{response_hist_compare}). In summary, there were 2,731 clips with both high Fluidity and Enjoyment, 92 clips with both low Fluidity and Enjoyment, 123 clips with high Enjoyment and low Fluidity, and 46 clips with low Enjoyment and high Fluidity. The contingency analysis showed related but not equivalent ratings for Enjoyment and Fluidity (\textit{$\chi^2$} = 758.13, \textit{p} $<$ 0.001).

\section{Modeling}

\subsection{Feature Extraction}

\subsubsection{Audio}
The audio embeddings were extracted using the popular state-of-art pretrained audio DNNs, including VGGish \cite{vggish}, YAMNet \cite{yamnet} and Wav2Vec2 (base-960h) \cite{baevski2020wav2vec}. VGGish and YAMNet are domain-general audio DNNs, while Wav2Vec2 is a speech-specific DNN. The group audios were first downsampled to 16 kHz, and then the VGGish, YAMNet, and Wav2Vec2 models return 128, 1024, and 768 embeddings per 0.96, 0.48, and 1.00 s, respectively.

\subsubsection{Facial Action}
The RoomReader corpus provided facial action unit intensity time-series per participant (17 features, i.e., blink and jaw drop. However lip suck data was not included), which was converted from each participant’s video data using \texttt{OpenFace} \cite{OpenFace7477553}. The original sample rate was 60 Hz, and we downsampled by window-averaging every .98 seconds. We averaged the time-series of each action unit across participants for each clip, resulting in 17 feature time-series per clip.

\subsubsection{Body Motion}
As our previous studies showed, the Granger causality (GC) coupling strength of body motion time series among in-personally interacting individuals can positively reflect their coordination outcomes in various social contexts (e.g., speed dating conversations). Therefore, the GC coupling among videoconference participants for each clip was calculated using our previous approach with a few modifications \cite{chang_body_sway_music, chang_body_sway_dating, chang2019body, wood2022creating}: (1) The distance time-series between each participant’s and webcam was quantified (sample rate 59--60 Hz), based on the \texttt{mediapipe} library \cite{lugaresi2019mediapipe}. (2) Each time-series was downsampled by window-averaging to 8 Hz, transformed into its first derivative to ensure stationary, and then z-normalized across time. (3) An averaged GC values among all pairwise participants was calculated by \texttt{GCCA()} function of the \texttt{MVGC} toolbox \cite{barnett2014mvgc} with model order 12 (1.5 s). 

\subsection{Model Training}
The machine learning model was trained using the \texttt{scikit-learn} (1.3) library in \texttt{Python} (3.8) with the following pipeline. The cross-validation included these preprocessing steps: imputing missing data with the median (or mode for event classification), z-normalizing each feature, and applying PCA for dimensionality reduction. We trained a logistic regression model using stochastic gradient descent (\texttt{SGDClassifier()}) with balanced class weights. We also experimented with other supervised learning models, including support vector machines, random forests, lightGBM, perceptrons, and more (not reported). However, we chose to report and interpret only the results from logistic regression due to its simplicity, robustness, and empirical performance in this study. We used \texttt{BayesianOptimization} \cite{bayesian_optimization_tool} for hyperparameter tuning (PCA explained variance: 50--99\%; alpha: 10$^{-10}$--1; elastic net penalty L1 ratio: 0--1) over 600 iterations. In each iteration, we performed 5-fold random-shuffled stratified group cross-validation, maximizing the macro-averaged ROC-AUC. This procedure ensures that the model predicts new videoconference sessions with new participants without data leakage. We also report the macro-averaged precision (AP), balanced accuracy (BA), and macro-averaged F1 score. Computation was performed using New York University’s High Performance Computing resources.

\section{Analyses and Results}

\subsection{Predicting Fluidity and Enjoyment}

We first investigated which combination of features, extracted from the 0--7 seconds, resulted in the model achieving the best ROC-AUC (Table \ref{results7sTab}). The models trained on features combining VGGish audio embeddings, facial action, and body motion GC best predicted Fluidity and Enjoyment. Different feature domains contributing differently to performance: the audio features are most useful, adding facial features can still marginally improve model performance, while the contribution of body motion is minimal.

\vspace{-10pt} 
\begin{table}[bp]
\caption{Model Performances on 0--7 s features}
\vspace{-10pt} 
\begin{center}
\begin{tabular}{cccccc}
\toprule
\textbf{Target} &  \textbf{Feature} & \textbf{ROC-AUC} & \textbf{AP} & \textbf{F1} & \textbf{BA}\\
\midrule
Fluidity &         Face             &   .668           &  .963       &  .440       &  .625 \\
Fluidity &         Face+Body        &   .668           &  .963       &  .440       &  .620 \\
Fluidity &         VGGish           &   .805           &  .977       &  .547       &  .712 \\
Fluidity &         VGGish+Body      &   .806           &  .977       &  .543       &  .709 \\
Fluidity &         VGGish+Face      &   .814           &  .980       &  .584       &  .728 \\
Fluidity &         VGGish+Face+Body &   .815           &  .980       &  .584       &  .728 \\
\midrule
Enjoyment &        Face             &   .758           &  .985       &  .485       &  .683 \\
Enjoyment &        Face+Body        &   .757           &  .985       &  .479       &  .685 \\
Enjoyment &        VGGish           &   .859           &  .990       &  .593       &  .788 \\
Enjoyment &        VGGish+Body      &   .860           &  .990       &  .590       &  .782 \\
Enjoyment &        VGGish+Face      &   .873           &  .992       &  .573       &  .788 \\
Enjoyment &        VGGish+Face+Body &   .874           &  .992       &  .565       &  .784 \\
\bottomrule
\end{tabular}
\label{results7sTab}
\end{center}
\vspace{-10pt} 
\end{table}

These results were not biased by the arbitrary selection of the threshold. Even when set to 3, the model trained on the same feature combination still performed best in predicting Fluidity (ROC-AUC: .785) and Enjoyment (ROC-AUC: .836), with only a slight decrease of about .04 in ROC-AUC. The following results are all based on a threshold of 2.5.

We also compared models trained on different audio DNN embeddings. However, the models with the highest ROC-AUCs using YAMNet (Fluidity: .694, Enjoyment: .789) and Wav2Vec2 (Fluidity: .815, Enjoyment: .841) were lower than or roughly equivalent to the models using VGGish embeddings. This suggests that while both speech-specific and domain-general audio DNN models can capture low-level conversational outcomes (i.e., fluidity), domain-general audio DNN models better represent high-level conversational outcomes (i.e., enjoyment).

Although body motion GC has little contribution to model performance, it was statistically higher among high Fluidity clips (\textit{t}(2990) = 2.29, \textit{p} = .022), replicating previous studies. However, it did not differ between high and low Enjoyment clips (\textit{t}(2990) = 0.51, \textit{p} = .607). While it shows a statistical association, its potential to enhance model performance could be significantly increased by calculating it over minutes, as in previous studies, rather than the 7-second period used here.

\subsection{Prediction Based on the Pre-event Features}

We examined how powerful the features \textit{before} the conversational event in the clips are in predicting Fluidity or Enjoyment by running additional models but using only the features from the first 3 seconds, while body motion was excluded as it is too short to calculate GC (Table \ref{results3sTab}). Interestingly, the models involving audio features dropped by more than .11 in ROC-AUC, but the models based on facial actions alone only dropped by .03 or less. This suggests that while audio features can dynamically reflect fluctuating conversational outcomes before and after the conversational event, facial actions represent the more long-term state of the conversation (e.g., people were already too bored to chat).

\begin{table}
\caption{Model Performances on 0--3 s features}
\begin{center}
\begin{tabular}{cccccc}
\toprule
\textbf{Target} &  \textbf{Feature} & \textbf{ROC-AUC} & \textbf{AP} & \textbf{F1} & \textbf{BA}\\
\midrule
Fluidity &         Face             &   .645           &  .957       &  .442       &  .610 \\
Fluidity &         VGGish           &   .669           &  .958       &  .490       &  .623 \\
Fluidity &         VGGish+Face      &   .690           &  .964       &  .516       &  .631 \\
\midrule
Enjoyment &        Face             &   .727           &  .981       &  .450       &  .661 \\
Enjoyment &        VGGish           &   .716           &  .977       &  .492       &  .676 \\
Enjoyment &        VGGish+Face      &   .761           &  .982       &  .522       &  .714 \\
\bottomrule
\end{tabular}
\label{results3sTab}
\end{center}
\vspace{-10pt} 
\end{table}

\subsection{Generalizability Across Fluidity and Enjoyment Models}
As Fluidity and Enjoyment ratings were positively associated, we further explored whether their models relied on the same underlying signals, by applying the best 7-s Fluidity model to predict Enjoyment and vice versa. The results showed that the Fluidity model can predict Enjoyment (ROC-AUC: .833, AP: .976, F1: .478, BA: .739), and the Enjoyment model can also predict Fluidity (ROC-AUC: .736, AP: .951, F1: .485, BA: .662). The decreases in ROC-AUC were small compared to Table \ref{results7sTab}. However, the drop from Fluidity to Enjoyment (.041), almost half of that in the opposite direction (.079), implies that low fluidity in conversation is more likely to lead to low enjoyment than vice versa.

\subsection{Classifying Conversational Events}
\begin{figure}
\vspace{-5pt}
\centerline{\includegraphics[width=0.5\textwidth]{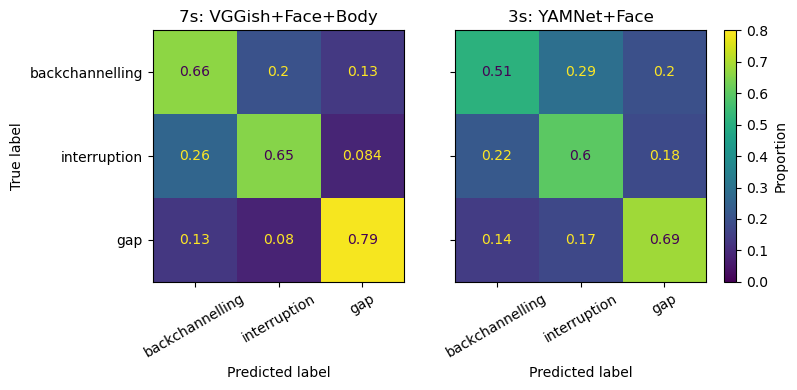}}
\vspace{-5pt}
\caption{Confusion Matrix of Classifying Conversational Events}
\label{Q2_cm}
\vspace{-10pt}
\end{figure}

We investigated whether conversational events can be differentiated. Among all combinations, the 7-s model trained on VGGish, facial, and body motion features performed the best (ROC-AUC: .867, AP: .773, F1: .696, BA: .697). The 3-s pre-event model trained on YAMNet and facial features performed the best (ROC-AUC: .778, AP: .640, F1: .593, BA: .599), though it was not as effective as the 7-s model, it could still differentiate the events. The confusion matrices showed that gaps and interruptions are distinguishable in both models, but classifying backchanneling relies more on post-event features (Fig. \ref{Q2_cm}). 

\section{Conclusions, Impacts, and Limitations}

The success of our model shows that, in videoconferencing, subjective high-level social outcomes and conversational events can be predicted by multimodal signals (audio, facial actions, body motion). This result will have a significant impact on the videoconference user experience research studies, by unlocking bulk analysis using a machine learning model rather than solely relying on expensive manual human annotation. In the future, a model such as this could be used by consumers, such as teachers, therapists and team managers, to identify specific awkward moments after a session, instead of relying on manual review or anecdotal memory. A realtime version of this model could potentially be used for in-the-moment interventions to facilitate more fluid turn-taking.

One limitation of this work is due to the RoomReader corpus: while its protocol is carefully designed to provoke naturalistic behavior, this likely makes the dataset overly self-consistent, and so the generalizability of our model to broader videoconferencing contexts remain untested. 

\section*{Acknowledgment}

We thank Iran R. Roman and our colleagues at New York University for their administrative and consultative support.

\bibliographystyle{IEEEtran}
\bibliography{derailing}

\end{document}